# COLOR-PHASE ANALYSIS FOR SINUSOIDAL STRUCTURED LIGHT IN RAPID RANGE IMAGING

*Changsoo Je*[1]    *Sang Wook Lee*[1]    *Rae-Hong Park*[2]

[1]Dept. of Media Technology    [2]Dept. of Electronic Engineering
Sogang University, Seoul, Korea
`{vision, slee, rhpark}@sogang.ac.kr`

**ABSTRACT**

Active range sensing using structured-light is the most accurate and reliable method for obtaining 3D information. However, most of the work has been limited to range sensing of static objects, and range sensing of dynamic (moving or deforming) objects has been investigated recently only by a few researchers. Sinusoidal structured-light is one of the well-known optical methods for 3D measurement. In this paper, we present a novel method for rapid high-resolution range imaging using color sinusoidal pattern. We consider the real-world problem of nonlinearity and color-band crosstalk in the color light projector and color camera, and present methods for accurate recovery of color-phase. For high-resolution ranging, we use high-frequency patterns and describe new unwrapping algorithms for reliable range recovery. The experimental results demonstrate the effectiveness of our methods.

## 1. INTRODUCTION

The goal of our research presented in this paper is to develop a realistic way of acquiring high-resolution dynamic (moving or deforming) 3D range data using a single video frame for real-time capture. 3D reconstruction of dynamic objects is highly challenging in that only a single projection of structured light pattern should be used for real-time capture.

Various active range-sensing systems have been suggested for high-resolution 3D ranging and recently for high-speed scanning. The laser stripe-based systems can estimate depths in high resolution because of the spatial sharpness of coherent light beam. However, they require mechanical motion of the laser-emitting device for object scanning, and this prohibits real-time sensing in most cases. Therefore, real-time sensing without mechanical scanning has to rely on the projection of specially designed incoherent structured light. The well-established structured-light sensing uses multiple projections of binary-encoded light stripe patterns, but this requirement of multiple projections also makes real-time imaging difficult. The depth resolution is dependent on the number of binary stripe patterns, i.e., the number of projections. A new BW stripe pattern method for real-time ranging was proposed [6], but it is appropriate only for slowly moving rigid bodies but not for deforming objects.

Color can be used to decrease the number of projections, and a color-encoded discrete-stripe structured light method has been suggested [1], and an attempt was made to minimize the number of projections of color patterns [4]. On the other hand, there have been approaches based on continuous color patterns such as rainbow and sinusoidal patterns. The rainbow range finder (RRF) uses color spectrum as a projection pattern for real-time scanning [3]. Among the continuous patterns, it may be noted that a method has been suggested for using the ratio of continuous gray-level patterns [5]. The phase-shifted three-cosine continuous pattern has been designed [2], and this uses phase of the three-sinusoid light instead of color spectrum.

The advantage of using color rather than gray levels is to increase the number of distinguishable patterns by three times when three RGB colors can be independently used. The advantage of using continuous patterns rather than discrete patterns is to increase the number of stripe ids (identification labels) per projection. On the other hand, color and continuous pattern have a number of problems for practical use in terms of color-band correlation, non-linearity and sensitivities to surface reflectance and noise. In this paper, we present our efforts to overcome the problems that arise when continuous color sinusoidal patterns are used for range imaging.

The sinusoidal structured-light is one of continuous patterns which can be easily applied to three color-channels for a 3-phase color pattern. The most important issue is the recovery of accurate color-phase. Color-phase recovery is disturbed by surface reflectance, color-band crosstalk, nonlinearity, noise, etc. In addition to color and photometric calibration, high frequency light patterns can reduce the influence of system nonlinearity and noise for higher resolution, and they can be achieved by using multiple periods of sinusoids. To increase depth resolution and decrease its sensitivity to noise and nonlinearity, multiple periods of high-frequency sinusoidal patterns are usually used and some appropriate unwrapping method is needed to disambiguate multiple appearance of color phase in a scene.

In this paper, we present our research on rapid high-resolution range imaging based on color sinusoidal structured-light. The focus of our research is on the investigation of accurate and efficient color-phase recovery, and reliable phase-unwrapping. The phase-shifted three-cosine fringe pattern has been suggested for real-time ranging [2]. However, the methods to overcome system nonlinearity and color-crosstalk, and unwrapping techniques for repetitive high-frequency fringe patterns have not been fully investigated.

The rest of this paper is organized as follows. Section 2 describes the design of three-color sinusoidal pattern for accurate 3D data acquisition, and section 3 presents methods for accurate color-phase recovery. Section 4 presents phase-unwrapping algorithm to remove discontinuity due to the repetitive color-phase. In section 5, experimental results are presented, and section 6 concludes this paper.

## 2. SINUSOIDAL PATTERN SYNTHESIS

In this section, we describe the sinusoidal structured-light method and consider projector-camera response. Optimal color sinusoidal projection pattern image is designed by compensating projector-camera response.

### 2.1. Sinusoidal Structured-Light Method

The phase-shifting method has been known for 3D measurement [7] [8], and Huang et al. applied three-phase sinusoidal pattern on 3 color-channels. Their intensities (RGB) are defined by the following equations [2]:

$$C_1(x,y) = a(x,y) + b(x,y)\cos[\phi(x,y) - 2\pi/3], \quad (2.1)$$

$$C_2(x,y) = a(x,y) + b(x,y)\cos[\phi(x,y)], \quad (2.2)$$

$$C_3(x,y) = a(x,y) + b(x,y)\cos[\phi(x,y) + 2\pi/3], \quad (2.3)$$

where $(C_1, C_2, C_3) \equiv (R, G, B)$, $a(x,y)$ is the average intensity, $b(x,y)$ is the intensity modulation, and $\phi(x,y)$ is the phase to be determined. Figure 2.1 shows the sampled intensities of three-color sinusoidal pattern. The phase of these cosine patterns is given by:

$$\phi(x,y) = \arctan\left(\sqrt{3}\frac{C_1 - C_3}{2C_2 - C_1 - C_3}\right). \quad (2.4)$$

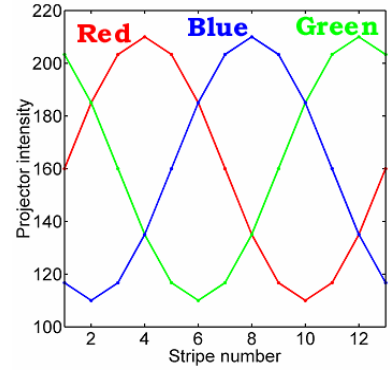

Figure 2.1 Three-phase sinusoids in RGB channels.

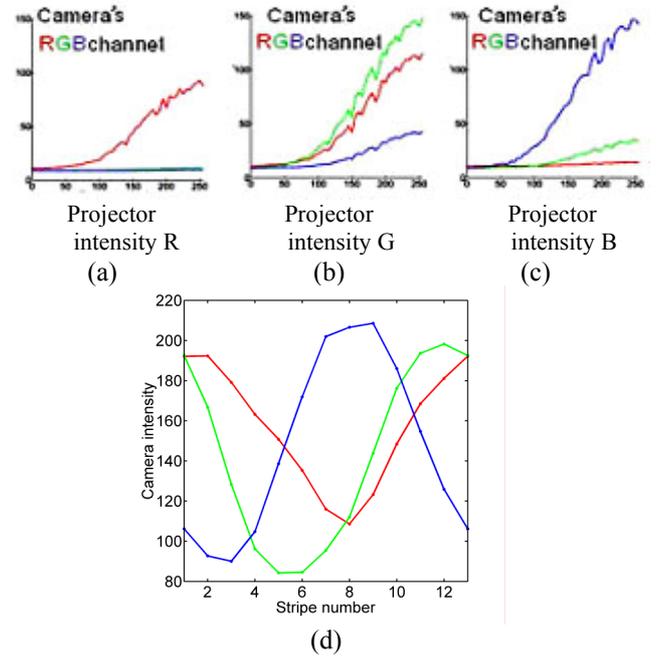

Figure 2.2 Camera responses to projector intensities: Camera's RGB-channel responses to projector's (a) R channel, (b) G channel and (c) B channel, and (d) three sinusoids measured by camera.

Determination of light planes (stripe ids or phases) with this continuous light pattern is highly dependent on the system linearity. We present simple processing methods which work almost as effectively as rigorous radiometric and sensor calibration.

### 2.2. Projector-Camera Response

Equations (2.1) ~ (2.3) assume that the camera response to projector is linear and RGB color-channels are uncorrelated, i.e.:

$$C_i^c = \alpha C_i^p + \beta, \quad i = 1, 2, 3, \quad (2.5)$$

where $C_i^c$ is the intensity in the image captured by the camera, $C_i^p$ is the intensity in the projection image, and $\alpha, \beta$ are constants. When the camera/projector system is linear and color channels are uncorrelated, the color phase can be obtained directly as follows.

$$\begin{aligned}\phi^c(x,y) &= \arctan\left(\sqrt{3}\frac{C_1^c - C_3^c}{2C_2^c - C_1^c - C_3^c}\right) \\ &= \arctan\left(\sqrt{3}\frac{(\alpha C_1^p + \beta) - (\alpha C_3^p + \beta)}{2(\alpha C_2^p + \beta) - (\alpha C_1^p + \beta) - (\alpha C_3^p + \beta)}\right) \quad (2.6) \\ &= \arctan\left(\sqrt{3}\frac{C_1^p - C_3^p}{2C_2^p - C_1^p - C_3^p}\right) \\ &= \phi^p(x,y).\end{aligned}$$

The projector-camera response is not linear in reality, and the problem of correlation between color bands significantly affects the results. Figure 2.2 (a), (b) and (c) show camera's R-, G- and, B-channel responses to the projector R channel, G channel and B channel, respectively. It can be seen in Figure 2.2 (a) that the projector R channel is responded nearly into the camera's R channel, but with substantial nonlinearity. However, Figure 2.2 (b) and (c) show significant crosstalk between the camera and projector color channels in addition to nonlinearity. Figure 2.2 (d) shows the three (distorted) sinusoids due to the nonlinearity and color-band cross-correlation. In principle, the projector-camera distortion can be compensated for by rigorous color calibration. However, we present simple filters that are almost as effective for sinusoidal-light based imaging. We compensate for color-band correlation by $3 \times 3$ filter, and nonlinearity is compensated for by adjusting color-phase distribution (section 3.2).

## 3. COLOR-PHASE RECOVERY

As shown above, the camera response to projector intensity is highly complicated (due to nonlinearity and color correlation), and it is difficult to determine the precise color-phase. We balance color towards the neutral color, and then obtain the initial color-phase from the balanced color. We also remove the nonlinearity by adjustment of color-phase distribution.

### 3.1. Color Balance

Equations (2.1) ~ (2.3) and (2.6) assume that the object reflectance (or albedo) is neutral. For non-neutral surfaces such as face where albedo varies gradually, we find that local color balancing "neutralizes" the color measurement and effectively reduces problems arising from non-uniform and non-neutral surfaces.

For local balancing, we first compute the average color vector in a certain size of area:

$$\overline{C}_i(x,y) = \frac{\sum_{x,y} C_i(x,y)}{\sum_{x,y}}, \quad (3.1)$$

and use it for normalizing the captured color as:

$$c_i = \frac{C_i}{\overline{C}_i}, i = 1, 2, 3. \quad (3.2)$$

The color-balanced color-phase is obtained by the equation,

$$\phi(\vec{c}) = ArcTan[\ \sqrt{3}(c_1 - c_3)\ ,\ (2c_2 - c_1 - c_3)\ ] \quad (3.3)$$

Using $c_i$ is especially effective for range sensing of human faces where albedo varies gradually. The averaging widow size can be larger for the uniform albedo such as that of a plaster figure.

### 3.2. Adjustment of Color-Phase Distribution

As mentioned earlier, the camera response to projector is non-linear. The nonlinearity can be compensated for using reference objects. However, in real-time ranging, the use of reference chart is very restrictive. Hence we remove the nonlinearity by the adjustment of color-phase distribution without any reference image based on the assumption that color phase is uniformly distributed. We project high-frequency structured-light pattern in which phase differential through the pixels is constant (See, for instance, Figure 2.3). For this case where many sinusoidal periods are present in the light projection, we may assume that the distribution of color-phase in the captured image is uniform. We sample the color phases in a captured image from the pixels with their intensities above a certain threshold to exclude low-intensity pixels, and equalize the color phases for their uniform distribution. We sample the pixels instead of using them all just to reduce the computational time. The processing method is described below.

First, we sort the sampled color-phases, and then partition the color-phases into $N$ bins such that each bin has the same number of color-phases. For $M$ color phases and $N$ bins, for instance, the number of phase values in each class is $M/N$. The adjusted phase value of a pixel included in the $i$-th bin is given as:

$$\phi_a(i) = (i-1)/N, \ 1 < i < N. \quad (3.4)$$

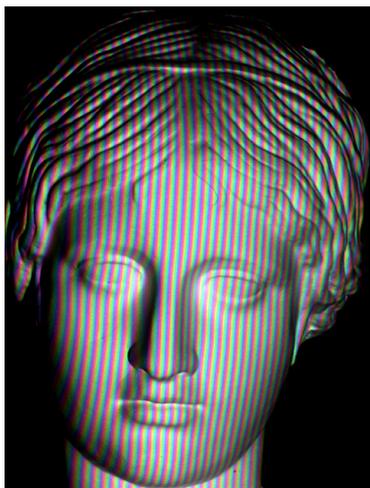

Figure 2.3 The captured image of Venus covered with the color sinusoidal pattern after compensation for color correlation.

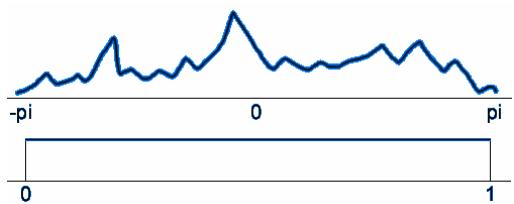

Figure 3.1 Color-phase distribution adjustment: the original distribution (top) and the adjusted distribution (bottom).

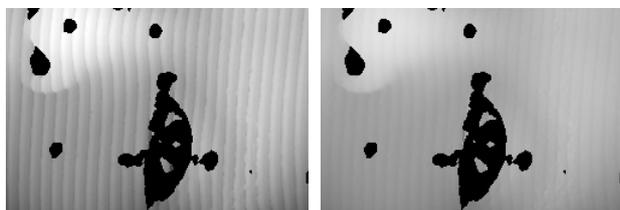

Figure 3.2 Color-phase distribution adjustment: range image before phase adjustment (left), and after phase adjustment (right).

The adjusted color phases have a "uniform distribution" as illustrated in Figure 3.1. Figure 3.2 shows the comparison of a range result by the original color-phase distribution with one by the adjusted distribution. We can see the decrease of ringing artifacts in the images.

## 4. PHASE-UNWRAPPING

The use of the multi-period pattern image results in repetition of the phase in the captured image, and therefore phase-unwrapping is needed. Unwrapping is to determine the number $n$ in the following equation:

$$\phi_u = \phi_a + 2n\pi . \tag{4.1}$$

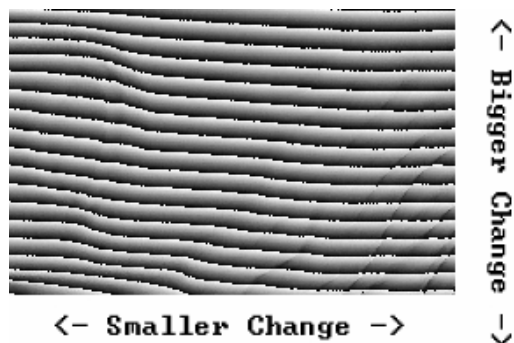

Figure 4.1 Directional phase change. The gray intensity levels in this image signify magnitudes of color-phases.

Our unwrapping method consists of initial phase-unwrapping and correction algorithms.

### 4.1. Initial Phase-Unwrapping Algorithm

Typically, the rate of color-phase change over neighboring pixels should be low, and it is different depending on the direction since the sinusoidal pattern we use is directional (see Figure 4.1). If the direction of the pattern is horizontal, the horizontal phase change is smaller than the vertical one, and *vice versa*. For the horizontal sinusoidal pattern shown in Figure 4.1, the smoothness constraint on the color phase can be more severe in the horizontal direction.

An optimization algorithm can be possibly used to determine the color phase at each pixel with the minimal phase difference rule between neighboring pixels as a smoothness constraint. However, constrained optimization is computationally expensive in general and we present simple filters for phase recovery.

The initial phase-unwrapping to determine the phase number $n$ is done with the following simple operator:

$$\phi_{u1}(x,y) = \phi_a(x,y) + round\left[\phi_{a\backslash u1}(neighbor(x,y)) - \phi_a(x,y)\right] \tag{4.2}$$

where $round[x]$ is the closest integer to $x$, and $\phi_a$ is the phase obtained from Equation (3.4). This operator initially starts from a pixel somewhere near the center of the image where its brightness is reliable, and recursively assigns proper phase numbers to the neighboring pixels according to the priorities of intensities, location and directionality until all the pixels are processed. For a pixel, the phase number is assigned only once. When the recursive process terminates before all the pixels are processed, a unprocessed pixel which has reliable priorities of intensities and location, and is neighboring a processed pixel, is taken as a new starting point.

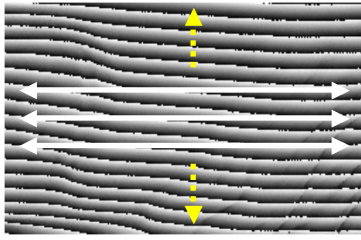

Figure 4.2  The direction of phase correction process. The gray intensity levels in this image signify magnitudes of color-phases.

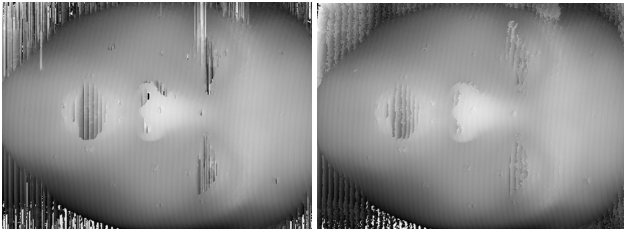

Figure 4.3  The range results: before (left) and after (right) the phase correction.

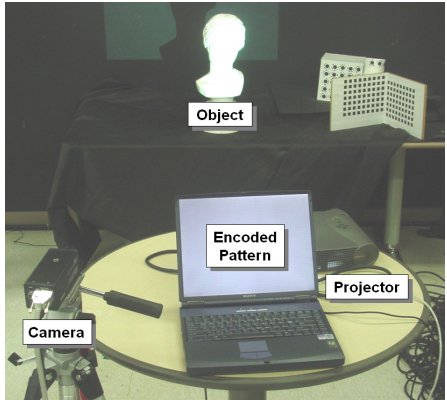

Figure 5.1  Experimental setup for capturing real-time range.

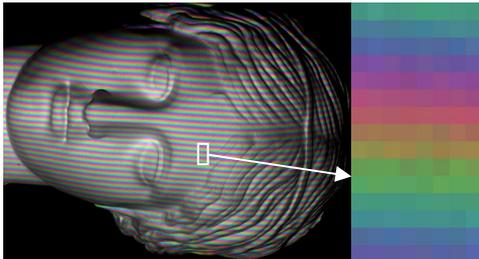

Figure 5.2  The captured image of Venus (left) and a local region is zoomed for viewing the projected color pattern (right).

This unwrapping algorithm assumes spatial smoothness of color-phase. When the smoothness assumption is violated in a few isolated regions, however, the algorithm works reasonably well as far as most of the pixels are connected due to the recursive propagation of minimal phase. Figure 4.3 (left) shows the result of initial unwrapping, and it can be seen that the phase numbers are recovered properly except in few regions. An additional filter is developed to correct the phase number errors in a few pixels.

### 4.2. Phase Correction Algorithm

The phase after the initial phase-unwrapping, $\phi_{u1}(x,y)$ is modified with the following filter:

$$\phi_{u2}(x,y) = \phi_{u1}(x,y) + \text{round}\left[\frac{\sum_{i,j}\phi_{u1}(x+i,y+j)}{\sum_{i,j}} - \phi_{u1}(x,y)\right]. \quad (4.3)$$

This filter minimizes sharp phase discontinuities using the mean phase from the surrounding window. The order in the processing is crucial. It starts from the center of the image and proceed in the minimal phase change direction first. In the example shown in Figure 4.2, the processing starts from the center and proceeds in the horizontal directions first (left direction first and then right direction) and then in the vertical directions (up first and then down). Despite its simplicity, this filter maximally utilizes the directional smoothness constraints. An example result of this processing is shown in Figure 4.3 (right).

## 5. EXPERIMENTAL RESULTS

Our experimental system consists of a Sony XC-003 3-CCD color camera ($640 \times 480$), an InFocus $1024 \times 768$ DLP projector and an Intel Pentium III PC (see Figure 5.1). We recovered the range data of a Venus plaster figure and a human face in a single video frame using the compensated (adjusted) color sinusoidal light pattern discussed earlier. We used a simple mean filter for smoothing and brightness thresholding for rejecting unreliable pixels for final processing. The window size for the phase correction filter was $(11 \times 11)$.

One cycle of color-phase in the captured image extends over about 12 pixels vertically (see Figure 5.2) and the whole image has 40 cycles of color-phase vertically. Since one cycle of color-phase is $2\pi$, the average phase differential through the pixel in vertical direction is about $\pi/6$. This sinusoidal pattern makes a distinct phase transition over vertically neighboring pixels.

Figure 5.3 (left) shows the VRML view of the Venus range data, and 5.3 (right) shows a rendered view after meshing the data. A result from a human face image is shown in Figure 5.4 (right). Note that the discontinuity of colors due to the painted markers for motion capture does not pose noticeable problems for phase unwrapping and correction.

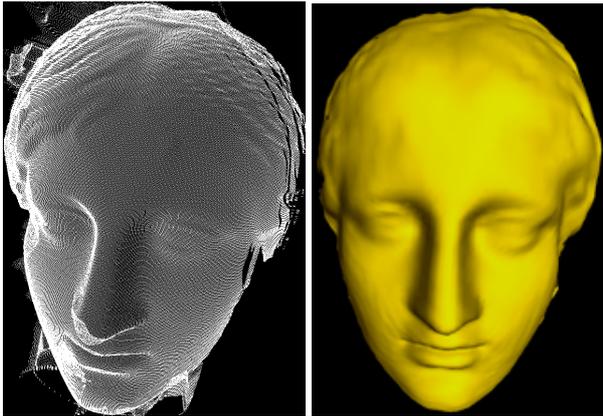

Figure 5.3 The VRML views of 3D points from Venus (left) and its rendered figure (right). This result is obtained from a single image frame.

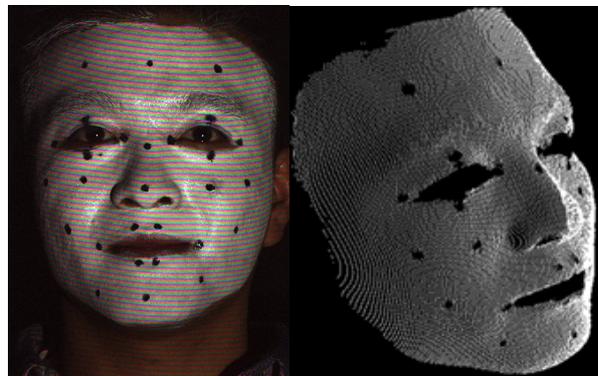

Figure 5.4 A human face scene image (left) and the VRML view of its 3D point data (right). This result is obtained from a single image frame.

## 6. CONCLUSION

This paper presents data analysis and processing methods for one-video-frame range sensing using phase-encoded color sinusoidal light. The presented algorithms are simple but highly effective for correcting color-phase distortion and unwrapping color-phase. Experimental results demonstrate the efficacy of the proposed methods for real-time range sensing.

The presented real-time range sensing method can be applied to many objects for dynamic modeling such as human facial expression and waving clothes. For future research, we are interested in applying the presented method to facial expression and cloth modeling. We are also interested in using multiple cameras and camera calibration for widening the field of view. Other research issues include: (1) investigation of more effective color patterns and theoretical foundations for more reliable unwrapping, (2) careful photometric calibration of projector and camera for lower distortion.

## 7. ACKNOWLEDGEMNETS

This work was supported by grant number R01-2002-000-00472-0 from the Basic Research program of Korea Science & Engineering Foundation (KOSEF), and also in part by the Sogang University Faculty Research Grant, 2003-2004. We would like to thank Mr. Jaesung Jo for his time and patience as the subject for facial range capture.